\documentclass[letterpaper]{article}
\usepackage{iccc}

\usepackage{times}
\usepackage{helvet}
\usepackage{courier}

\usepackage{graphicx}
\usepackage{hyperref}

\title{Transfer learning for Underrepresented Music Generation}

\author{Anahita Doosti \and Matthew Guzdial\\
Department of Computing Science, Alberta Machine Intelligence Institute (Amii)\\
University of Alberta\\
Edmonton, AB, Canada\\
\{doostisa, guzdial\}@ualberta.ca\\
}

\begin{document} 
\maketitle
\begin{abstract}
\begin{quote}
This paper investigates a combinational creativity approach to transfer learning to improve the performance of deep neural network-based models for music generation on out-of-distribution (OOD) genres. We identify Iranian folk music as an example of such an OOD genre for MusicVAE, a large generative music model. We find that a combinational creativity transfer learning approach can efficiently adapt MusicVAE to an Iranian folk music dataset, indicating potential for generating underrepresented music genres in the future.
\end{quote}
\end{abstract}

\section{Introduction}

Automated music generation has a long history \cite{briot2017deepSurvey}. In recent years, large-scale neural network models for music generation have arisen, trained on massive datasets and requiring significant computation \cite{CIVIT2022118190}. 
While these approaches have proven successful at replicating genres of music like those in their training sets, due to the nature of large-scale neural network models we expect this may not prove true for dissimilar genres. 
Specifically, we hypothesize that these large scale models will perform poorly for out-of-distribution (OOD) genres of music, those representing underrepresented or less globally popular types of music. 
We therefore conducted a study on one such large-scale neural network model to understand (1) how it performed on OOD music genres, and (2) how we might best adapt the model to an OOD music genre.

For this paper, we focus on Google Magenta's MusicVAE model \cite{musicvae2018}. We lack the space for a full discussion of the model, but direct interested readers to the original paper. This hierarchical variational autoencoder is trained on an enormous dataset of roughly 1.5 million unique MIDI files collected from the web. While its exact dataset was not made public, online repositories of MIDI files are typically made up of fan-made annotations of popular songs. Thus, automatic and indiscriminate data collection would result in an unbalanced dataset in terms of genre diversity. This is due to the fact that popular chart-topping songs are much more likely to be annotated in the MIDI format. 
The training requirements for MusicVAE are a problem when it comes to generating underrepresented music, like experimental music or music from particular cultures with distinct musical traditions.
These genres of music are unlikely to have the massive datasets needed to train models like MusicVAE. 
Even if such datasets existed, new musical genres are constantly being invented, meaning we could never use this approach to generate all underrepresented genres of music.

%
If we want to be able to generate underrepresented music, one option outside of training MusicVAE from scratch is transfer learning \cite{tan2018survey}. Transfer learning refers to the collection of approaches that can adapt knowledge from a model pre-trained on some source dataset (i.e., popular MIDI files) to a target domain with limited data (i.e., underrepresented music MIDI files). 
However, these approaches tend to require significant similarity between the source and target domains, which may not hold true for popular and underrepresented music genres \cite{marchetti2021convolutional}. 
Combinational creativity, also sometimes combinatorial creativity, is a type of creative problem solving in which two conceptual spaces are combined to represent a third or new conceptual space \cite{boden2009computer}. 
While different musical genres may vary in terms of their local features (e.g., melodies), they are all still music. 
As such, we hypothesized that a combinational creativity-inspired transfer learning approach may be able to outperform traditional transfer learning approaches at the task of adapting MusicVAE to an underrepresented genre of music \cite{Mahajan2023}.


In this paper, we explore the application of CE-MCTS \cite{Mahajan2023}, a combinational creativity-based transfer learning approach to adapt MusicVAE to an underrepresented music genre.
While there are many deep neural network (DNN) models like MusicVAE for music generation, applying transfer learning to a DNN model for music generation remains under-explored \cite{svegliato2016deep,marchetti2021convolutional}.
In addition, while combinational-creativity-based transfer learning approaches have been applied to many domains including image classification \cite{banerjee2021combinets} and financial health prediction \cite{Mahajan2023}, they have never been applied to the music generation domain.
In the remainder of this paper, we first demonstrate an experiment to identify an out-of-distribution (OOD) music genre for MusicVAE. 
We then introduce CE-MCTS and a number of more standard transfer learning baselines. 
Finally, we demonstrate their performance in terms of reconstruction accuracy for an OOD music dataset and present a short discussion on their music generation performance. 

\subsection{Related Work}

There have been many recent applications of deep neural networks (DNN) to music generation \cite{CIVIT2022118190}. For instance, sequence-based approaches are popular in this field due to their ability to learn long-term dependencies in musical pieces. Multiple studies have combined sequence-based models such as Long Short-term Memory (LSTM) Recurrent Neural Networks (RNN) with autoencoders and achieved good results \cite{perfrnn2017}. Alternatively, Generative Adversarial Networks (GAN) have been employed to generate novel music \cite{yang2017midinet}. 
These models typically are also trained from scratch. However, given  the data imbalance across different genres, approaches like transfer learning that can adapt knowledge from one domain to another might be useful. One such example attempts to test a pre-trained Generative Adversarial Networks (BinaryMuseGAN) with traditional Scottish music and improves its performance using finetuning \cite{marchetti2021convolutional}. To the best of our knowledge, finetuning is the only transfer learning approach that has been applied to DNN music generation \cite{svegliato2016deep}. 
We use it as a baseline.

In regards to Iranian (Persian) traditional or folk music, the prior work focuses on generation of music via traditional non-machine learning approaches and/or training from scratch. In \cite{SaharPhD}, the author uses a combination of evolutionary algorithms, Boltzmann machine models and cellular automata to generate music. They evaluate their work by the use of surveys targeted to both general and professional audiences. Alternatively, researchers have employed RNNs trained on a dataset of traditional Iranian music to generate music \cite{rnn-persian}. We note our goal is not to generate Iranian music specifically, but to explore the best ways to adapt large DNN music generation models to OOD genres like Iranian music. 

\subsection{Genre Analysis}
MusicVAE as a music generation model boasts a very impressive performance. Specifically, it achieves 95.1\% over its test dataset. However, our hypothesis was that MusicVAE would do poorly for OOD music. 

To examine this question, we collected four experimental datasets of 10 songs each. These were small in size as we only required a general approximation of whether the genre was out-of-distribution for MusicVAE.
We selected these songs according to two criteria: (1) if they were published after MusicVAE or were otherwise unlikely to be included in the original dataset \cite{musicvae2018} and (2) if their genres were distinct from a melodic standpoint. Melodies can differ in many ways such as contour, range and scale and these characteristics are different across different genres \cite{melodytheory}.

Our four datasets are as follows:
\begin{itemize}
    \item \textbf{Synth pop}, songs from a 2021 Netflix special, Inside by Bo Burnham, which musically fall into the synth pop category. This dataset serves as a comparison point, since we expected MusicVAE to perform well on this genre.
    \item \textbf{Iranian folk}, arose from a region with a long-standing history of composing music with independent roots from western music. As such, we anticipated this would be the most challenging for MusicVAE. 
    \item \textbf{Video game}, consists of Nintendo Entertainment System video game music. These songs have limited polyphony as only 3 notes can be played on the NES at once. 
    \item \textbf{Horror scores}, were designed to build suspense and create a sense of foreboding. Musically this genre frequently uses dissonant notes or chords, atonality (not having a clear scale), sudden changes of tempo, and other effects to induce a sense of eeriness and dread. 
\end{itemize}

\begin{table}[tb]
    \centering
    \begin{tabular}{|c|c|}
    \hline
    Dataset & Accuracy(\%) \\
    \hline\hline
    Synth pop & 95.83 \\
    \hline
    Iranian folk & 43.75 \\
    \hline
    Video game & 84.38 \\
    \hline
    Horror score & 87.92 \\
    \hline
    \end{tabular}
    \caption{MusicVAE accuracy on 4 datasets of different genres}
    \label{table:4Genre}
\end{table}

In these experiments, we fed melody sequences extracted from the songs into the pre-trained MusicVAE and report the reconstruction accuracy.
We include the results of our analysis in Table \ref{table:4Genre}.
As we expected, the model performed best on the first dataset.
The 95.83\% accuracy is in line with what was reported for the test accuracy on the original MusicVAE dataset. 
Predictably, the accuracy is noticeably lower for the other three datasets, with the Iranian dataset standing at a mere 43\%, a major drop in performance compared to the rest. 
As such, we focused on Iranian folk music for the remainder of our study.

\section{Iranian Folk Music Dataset} \label{data}   
We gathered a new dataset of Iranian folk MIDI files in order to evaluate the possibility of adapting MusicVAE to this out-of-distribution (OOD) genre.
This dataset consists of 100 MIDI files from both Farsi-speaking websites and from \url{musescore.com} which is a free sheet music sharing website. These files contain different instruments and varying levels of polyphony. 
We collected 100 MIDI files as this is in line with the target genre dataset size for prior finetuning-based transfer learning approaches with music generation models \cite{svegliato2016deep,marchetti2021convolutional}. 
However, we anticipate that CE-MCTS could perform well with fewer samples \cite{Mahajan2023}.
During our experiments we used a five-fold cross validation, meaning we split the data into five train-test splits (80 songs for training, 20 for testing), which helped ensure that we did not just get a ``lucky'' train-test split.

\section{Conceptual Expansion Monte Carlo Tree Search (CE-MCTS)}

We hypothesized that a combinational creativity-based transfer learning approach could most effectively adapt MusicVAE to an OOD genre. 
While there have been several prior examples of combinational creativity-based transfer learning approaches \cite{banerjee2021combinets,singamsetti2021conceptual}, we chose Conceptual Expansion Monte Carlo Tree Search (CE-MCTS)~\cite{Mahajan2023}. 
None of these combinational creativity-based transfer learning approaches have been applied to music data, but CE-MCTS demonstrated the ability to adapt to the behaviours of distinct groups of humans across problem domains. 
As such, we anticipated it would be the best for adapting to other types of human expression, like music.

Here we briefly describe CE-MCTS and how we adapted it in this work. However, for a full description of the approach we direct interested readers to \cite{Mahajan2023}.
The ``Conceptual Expansion'' (CE) in the name refers to the fact that we are searching over combinations \cite{banerjee2021combinets}. 
In this case, different combinations of the learned features from the original MusicVAE model. 
While it may seem unintuitive to combine features from the same model, this can allow us to approximate unseen features.
This is equivalent to combining different musical patterns in the source domain (i.e., popular MIDI music) to approximate patterns in the target domain (i.e., Iranian folk music).
We then employ Monte Carlo Tree Search (MCTS) to search over the space of these possible combinations. 
As is typical with MCTS, we build up a tree to search through this space. 
The root node represents the original trained MusicVAE model (no combinations) and every subsequent node represents a different set of feature combinations, with closer nodes representing more similar combinations. 



For our implementation of CE-MCTS we largely employed the same setup from the original paper \cite{Mahajan2023}. 
However, we made a number of changes for this domain.
For the fitness, we used the reconstruction accuracy of the training split (80 songs from Iranian folk music dataset).
We ran 10 iterations, each with 10 rollouts ($L=5$) and $\epsilon=0.5$.
We did this to bias the search towards exploration near the original MusicVAE model as we did not want to risk catastrophic forgetting, in which a model loses useful features.
Ultimately, we output the three best performing models according to the fitness and report their average performance over the test split (20 songs). 
The strategy for the final model selection varies based by domain.

\section{Transfer Learning Baselines} \label{tlmethod}

We hypothesized that CE-MCTS would outperform standard transfer learning approaches at adapting MusicVAE to our Iranian folk music dataset. Here we introduce the transfer learning baselines we used for comparison purposes. Other transfer learning approaches were not appropriate as we lacked access to MusicVAE's original training dataset.

\begin{itemize}
    \item \textbf{Non-transfer}, where we train a randomly initialized MusicVAE on the Iranian music dataset alone. This represents the standard approach to this problem without transfer learning. We did not expect this to work given the limited amount of training data.
    \item \textbf{Zero-shot}, which uses the pre-trained weights of MusicVAE with no additional training on the Iranian music dataset. We know MusicVAE does poorly when reconstructing 10 Iranian folk songs, but this won't necessarily hold for our larger 100 song dataset.
    \item \textbf{Finetuning (all)}, in which we use finetuning, a traditional network-based transfer learning approach that has been applied in prior music generation DNN transfer learning work \cite{tan2018survey,svegliato2016deep,marchetti2021convolutional}. In this baseline, we applied finetuning by continuing to train MusicVAE on a train split of our dataset until convergence. The (all) indicates retraining all of MusicVAE's layers. This is unusual as it can lead to catastrophic forgetting, where useful features are lost in the adaptation process.
    \item \textbf{Finetuning (last)}, which is the same as Finetuning (all) except that we freeze the weights of all but the last layer. This is the more typical approach when applying finetuning as it assumes that the earlier layers contain useful features (e.g., musical patterns) and we can just adapt the last layer to apply these patterns more appropriately for our target dataset.
\end{itemize}

We also developed a knowledge distillation approach called student-teacher learning \cite{student-teacher}. In this approach we trained a MusicVAE (student network) on Iranian music, through a combination of its loss and the loss of another MusicVAE with pre-trained weights (teacher network). This model proved worse than all other baselines, thus we do not include its results.

\section{Evaluation}

For our evaluation, we compared the reconstruction accuracy of our approach and baselines. 
Each model was fed our test set and we measured the percentage of correctly reconstructed notes.
Clearly, this method of evaluation does not assess the ability of the model to actually generate new musical sequences, which is the main objective of a music generation model. 
However, this approach allows us to run an initial quantitative evaluation. 
While there are objective metrics employed by other researchers, they are not consistently defined, making it difficult to compare outputs across different generation systems. Furthermore, there is no correlation between qualitative and quantitative metrics of evaluation, making it difficult to draw implications. We expect this to be even more difficult for a genre like Iranian music or other regional genres. Therefore, correctly evaluating the quality of generation would require a human subject study with experts in the target genre (i.e. Iranian folk music).~\cite{ji2020comprehensive} 
We leave this to future work, but include a small case study below.

\section{Results}

\begin{table}[tb]
    \centering
    \scalebox{0.75}{
    \begin{tabular}{|c||c|c|c|c|c|c|}
        \hline
        Approach & Fold 1 & Fold 2 & Fold 3 & Fold 4 & Fold 5 & Average  \\
        \hline\hline
        Non-transfer & 68.75 & 68.75 & 68.75 & 68.75 & 68.75 & 68.75 \\
        
        \hline
        Zero-shot & 93.75 & 90.62 & 84.37 & 87.50 & 87.50 & 88.75 \\
        \hline
        Finetune (all) & 87.50 & 84.37 & 78.12 & 78.12 & 75.00 & 80.62  \\
        \hline
        Finetune (last) & 96.87 & 90.62 & 93.75 & \textbf{100} & \textbf{100} & 96.25 \\
        \hline
        CE-MCTS & \textbf{98.96} & \textbf{94.80} & \textbf{98.97} & 94.84 & \textbf{100} & \textbf{97.52} \\
        \hline
    \end{tabular}
    }
    \caption{Training reconstruction accuracy of each approach on each fold}
    \label{table:train_res}
\end{table}

\begin{table}[h]
    \centering
    \scalebox{0.75}{
    \begin{tabular}{|c||c|c|c|c|c|c|}
        \hline
        Approach & Fold 1 & Fold 2 & Fold 3 & Fold 4 & Fold 5 & Average  \\
        \hline\hline
        Non-transfer & 37.50 & 37.00 & 37.50 & 6.25 & 53.12 & 34.27 \\
        \hline
        Zero-shot & 90.62 & 87.50 & 75.00 & 37.50 & 90.62 & 76.24 \\
        \hline
        Finetune (all) & 81.25 & 68.75 & 25.00 & 34.37 & 65.62 & 55.00 \\
        \hline
        Finetune (last) & \textbf{96.87} & 96.87 & 65.62 & 40.62 & 93.75 & 78.75\\
        \hline
        CE-MCTS & 93.75 & \textbf{97.9} & \textbf{83.34} & \textbf{51.07} & \textbf{93.77} & \textbf{83.97}\\
        \hline
    \end{tabular}
    }
    \caption{Test reconstruction accuracy of each approach on each fold}
    \label{table:test_res}
\end{table}

Tables \ref{table:train_res} and \ref{table:test_res} contain our training and test reconstruction accuracy results, respectively. 
Overall we can observe that CE-MCTS consistently outperforms other methods in both training and test accuracy. Although Finetuning (last) performs similarly on average during training, CE-MCTS is better at reconstructing the test data. 

As we expected, Finetuning (all) produces inferior results to Finetuning (last). In fact, it seems that the original pre-trained MusicVAE (Zero-shot) outperforms this method. This is likely due to catastrophic forgetting. As for the non-transfer method, it is not surprising that the small quantity of data available is unable to effectively train the network. The initial dataset size used to train MusicVAE is roughly 15,000 times larger than our dataset.

CE-MCTS outperforms both the pre-trained MusicVAE and last layer finetuning on test accuracy. Therefore we can deduce that by recombining the features in the earlier layers, CE-MCTS is able to create better features for the target dataset. Based on the performance of the Zero-shot and Finetuning (last) baselines, we can infer that the features present in original model are not sufficient to represent Iranian folk music.
This suggests we can usefully approximate Iranian folk music features via a combination of features from popular western music.

\section{Case Study} \label{qualitative}
In this section, we provide a brief qualitative analysis of some of the models in terms of music generation instead of reconstruction. We include figures with representative outputs from the three best performing approaches. These examples were chosen by the first author, who has expertise in Iranian music, to be generally representative of the characteristics of the outputs from these approaches. This is obviously highly subjective and susceptible to confirmation bias. A study with expert participants is needed to make reliable assertions about the quality of generation, but we leave this to future work.

In each corresponding figure, the x-axis represents time in seconds, limited to 4 seconds which is the length of all 2-bar outputs by MusicVAE. The y-axis represent the pitch for the notes in the MIDI format which ranges between 0 and 127. Each red rectangle in the figure represents a continuous note.

\begin{figure}[h]
    \centering
    \includegraphics[keepaspectratio=true, width=0.4\textwidth]{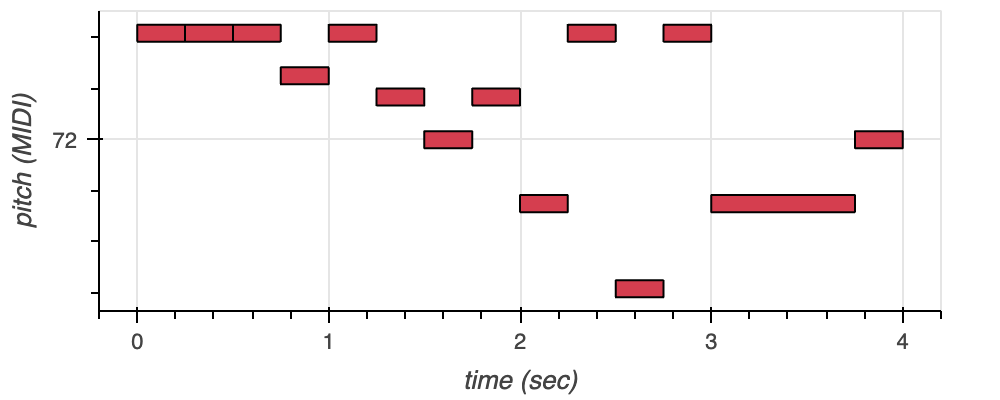}
    \caption{Visualization of a melody generated by the pre-trained MusicVAE model}
    \label{fig:musicvae}
\end{figure}

\begin{figure}[tbh]
    \centering
    \includegraphics[keepaspectratio=true, width=0.4\textwidth]{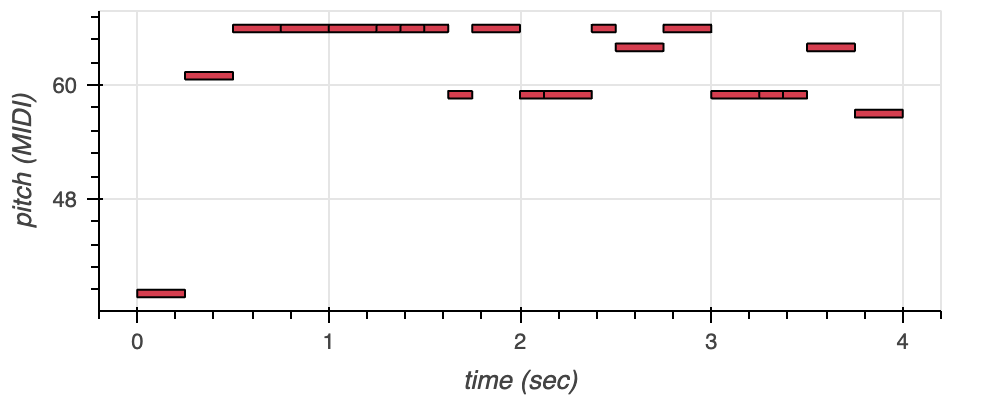}
    \caption{Visualization of a melody generated by the model finetuning the last layer}
    \label{fig:finetunelast}
\end{figure}

\begin{figure}[tbh]
    \centering
    \includegraphics[keepaspectratio=true, width=0.4\textwidth]{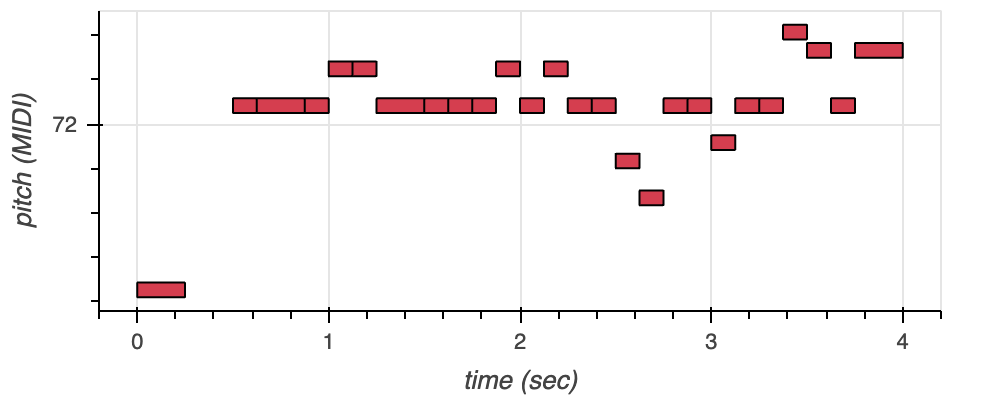}
    \caption{Visualization of a melody generated by the CE-MCTS model}
    \label{fig:cemcts}
\end{figure}

Figure \ref{fig:musicvae} represents a typical melody generated by the Zero-shot pre-trained MusicVAE. The notes in this melody sound harmonious and follow a somewhat cohesive progression. They also gradually move from a higher pitch to a lower one, spanning somewhat evenly across the melodic range (distance between the highest and the lowest pitch). 
Figures \ref{fig:finetunelast} and \ref{fig:cemcts}, were generated using the Finetuning (last) and CE-MCTS approaches, respectively. In the first author's subjective opinion, samples generated by these two models sound more similar to, and evocative of, the type of melodies present in Iranian folk music. This is hard to qualify but here we point out a number of characteristics commonly seen in traditional and folk Iranian (Persian) music according to \cite{farhat2004dastgah} and \cite{iranian_classical_music}. 
\begin{itemize}
    \item Melodies have a narrow register (pitch range).
    \item Melodic movement is often achieved with conjunct steps.
    \item There is an emphasis on cadence, symmetry, and repetition of musical motifs at varying pitches.
    \item Rhythmic patterns are generally kept uncomplicated and rhythmic changes are infrequent.
    \item The tempo is often fast, with dense ornamentation. Similar to this, it is common to see repetitive and rapid use of the same note/pitch.
\end{itemize}
As shown in both figures, the register is more limited locally and patterns that repeat the same note appear. 

\section{Conclusions}
In this paper, we investigated how MusicVAE, a music generation model, can be adapted to OOD music.
We identified that MusicVAE in particular struggles with Iranian folk music. We then explored different transfer learning methods in order to improve MusicVAE's performance on a newly collected Iranian folk music dataset. Based on our results, we observed that CE-MCTS, a combinational creativity-based transfer learning approach, is better able to produce reconstructions of this genre of music. This suggests that we can successfully adapt these large music generation models for underrepresented genres of music, and that combinational creativity can be an especially helpful tool in this task.






\bibliographystyle{iccc}
\bibliography{iccc}

\end{document}